\documentclass[conference]{IEEEtran}
\IEEEoverridecommandlockouts

\usepackage{cite}
\usepackage{amsmath,amssymb,amsfonts}
\usepackage{algorithmic}
\usepackage{graphicx}
\usepackage{textcomp}
\usepackage{xcolor}
\usepackage{mdframed}
\usepackage{multirow}
\usepackage{hhline}
\usepackage{hyperref}

\begin{document}
\title{Co-design Optimization for Underwater Vehicle Docking Systems}
\author{Jonathan Wallen$^{*}$, Maddyson Jeske$^{*}$, and Zhuoyuan Song 
\thanks {$^{*}$JW and MJ contributed equally to this work.}
\thanks {The authors are with the Department of Mechanical Engineering, University of Hawai`i at M\={a}noa, Honolulu, HI 96822, USA. E-mails: \tt\small \{jwallen, mjeske, zsong\}@hawaii.edu}
\thanks {Code is available at \url{https://github.com/song-ranlab/design-optimization}.}
}

\maketitle
\begin{abstract}
The design of autonomous underwater vehicles (AUVs) and their docking stations has been a popular research topic for several decades. Although many AUV and dock designs have been proposed, materialized, and commercialized, most of these existing designs prioritize the functionality of the AUV over the dock, or vise versa; there has been limited formal research in analytical optimization for AUV docking systems. In this paper, a multidisciplinary optimization framework is presented with the aim to fill this theoretical gap. We propose a co-design optimization method that optimizes multiple design parameters governing the archetype of an AUV and its docking system. Capturing the user design intents in the optimization process, the proposed method produces a set of optimal design parameters that satisfies a set of predefined bounds, constraints, and initial conditions. Three cases of design optimization are reported for different design intents. Each optimal design found in the three cases is compared to an existing system to show the validity of this design optimization framework.
\end{abstract}

 \section{Introduction}
\begin{figure}
    \centering
    \includegraphics[width=1\linewidth]{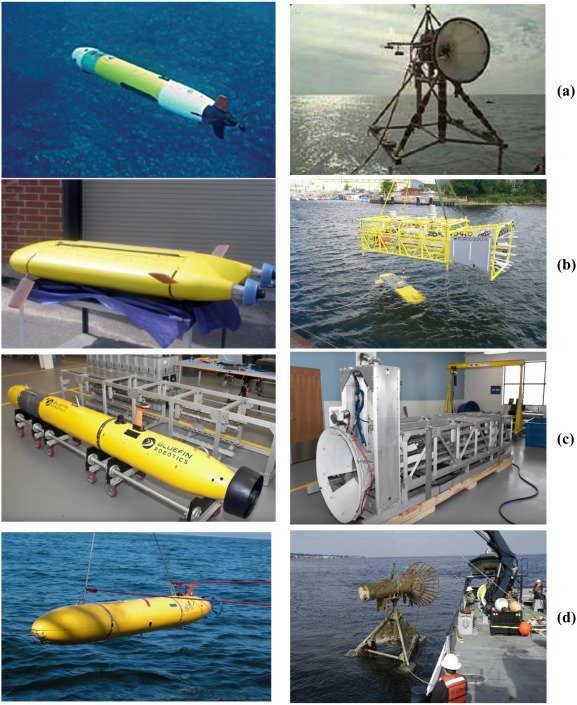}
    \caption{A survey of existing docking station designs conducted by Yazdani and co-authors in~\cite{YazdaniA:19}: (a) REMUS docking system; (b) Maridan Martin AUV with Eurodocker system; (c) Bluefin docking system; (d) Dorado docking system. The docking station design and geometry has been heavily influenced by the morphology of the AUV, and funnel-type docks are popular due to their design simplicity in serving slender-body AUVs.}
    \label{fig:1}
\end{figure}

Autonomous underwater vehicles (AUVs) have been widely used in the private and public sectors for more than six decades for a broad range of applications ranging from scientific exploration to defense~\cite{AntonelliG2008}. As the related technologies mature, an outstanding constraint for AUVs has persisted in their need for human intervention, particularly during their deployment and retrieval, which inflates their operational costs and diminishes their advantages over conventional platforms. Early efforts to dock an AUV to a source of power were conducted by Woods Hole Oceanographic Institute (WHOI) in the early 1990s~\cite{CurtinT:93a}. Since then, many entities have developed and tested AUV docking stations with varying degrees of success \cite{SinghH:01a,McEwenR:08a,YazdaniA:19,SongZ2020,PageB:21a}. Interestingly, many AUV docking stations consist of a guidance funnel to simplify the entry of slender-body AUVs while minimizing the design complexity of the dock. In \cite{YazdaniA:19}, a survey on the existing docking station designs was performed, revealing this pattern of a dominant archetype among existing designs (Fig.~\ref{fig:1}). In addition, the design and optimization of the docking station often emerges as an afterthought for an existing AUV platform. Recently, there have been a group of outlier designs as well. Two examples include a design resembling the geometry of airplane hanger \cite{AlbiezJ:15a} and a design similar to a fishing pole lure \cite{kimballP2018}. 

One key similarity among the investigations of funnel-type docks and other unique designs is that there is little to no formal investigation conducted or reported on the design optimization and modeling process. It can be inferred that much of the design work populating the area of AUV docking systems has been performed heuristically by experts with extensive empirical experiences with AUVs. However, recent studies found that both the AUV and dock's parameters can affect the overall success rate of a docking system~\cite{MartinS:16a,FletcherB:17a}. The goal of this paper is to fill the long-standing theoretical gap in the methodology for AUV-dock co-design by introducing a multidisciplinary design optimization (MDO) framework \cite{PapalambrosPanosY2017PoOD}. Rather than relying on subject-matter expertise alone, a formal approach is proposed that combines analytical modeling with optimization techniques to realize an optimally co-designed system.

The optimization framework presented here is characterized with the relevant physical models and interactions, design variables and their bounds \cite{shahv2007}, and the objective function along with constraints. To evaluate the proposed MDO method, we compare the results to several existing AUV docking station systems. This comparison shows the effectiveness of the co-design framework in arriving at design parameters consistent with the existing proven systems.

\section{Multidisciplinary Design Optimization Framework}

The intent of our MDO framework is captured by the following \textit{design statement}:

\smallskip
\noindent
\textit{To maximize the docking success and system versatility and minimize the hydrodynamic loss and cost of an autonomous underwater vehicle docking station system by varying vehicle and dock parameters while meeting constraints, bounds, and specific needs of the user/designer.}
\smallskip

An additional aim of this MDO methodology is to make it applicable to as many design variations as possible, meaning that we strive not to constrain the optimization solution space by adding design bias to the models and parameters. The proposed model is structured to receive user-defined optimization weights to suit their real-world design intent \cite{pamplambrosP2010}. Depending on the weight selection, an optimal set of design variables corresponding to a real-world implementation can be achieved. 

An example of such a design process is illustrated as follows:
\begin{enumerate}
    \item User determines the design intents of the AUV docking system and defines design parameters.
    \item User quantifies the design intents by identifying the optimization metrics and specifying their corresponding weights (i.e., importance).
    \item The optimization is performed based on the aforementioned user inputs with design parameters as the optimization variables.
    \item An array of optimized design parameters are obtained that govern the co-design of the AUV (e.g. geometry, mobility, etc.) and the docking station (e.g. geometry, ways-to-entry, etc.). 
\end{enumerate}


The proposed optimization framework is characterized with the relevant physical models and interactions, including design variables and their bounds, the objective function, and the applicable constraints. To construct the underlying MDO, each of the design objectives are weighted and combined into a total optimization cost function. 

A complete list of the design variables, objectives, and optimization parameters in the MDO problem for AUV-dock co-design are summarized in Table \ref{table:masterchart}. We cast the co-design optimization into the nonlinear programming (NLP) framework \cite{steuerr1986}. We define the optimization objective vector as
\begin{equation}
    \mathbf{j}=
    \begin{bmatrix}
    h,\ c,\ d,\ v 
    \end{bmatrix}
    ^T,
    \label{Eq.J}
\end{equation}
which is comprised of the system objectives that can be scaled by user specified weights. The corresponding weight vector (i.e., optimization parameters) can be defined as
\begin{equation}\label{eq:weight_vector}
    \mathbf{w}=
    \begin{bmatrix}
    p,\ q,\ -r,\ -s
    \end{bmatrix}.
\end{equation}
The design variables for the optimization make up the design vector 
\begin{equation}
    \mathbf{x}= 
    \begin{bmatrix}
    A,\ l,\ u,\ e,\ \eta 
    \end{bmatrix}
    ^T.
    \label{Eq.X}
\end{equation}
These design variables correspond to real-world properties that influence the systems of interest such as hydrodynamic drag, build cost, and control uncertainty of the AUV\cite{fossenT1994}. The design variables and objectives are explained in more detail in Sections \ref{des} and \ref{opt}, respectively. 
\begin{table}
\caption{Variables, objectives, and parameters for AUV-Dock co-design optimization}
\begin{center}
\begin{tabular}{|c|c|c|c|} 
 \hline
 \textbf{Symbol} & \textbf{Definition} & \textbf{Type} \\ 
 \hhline{|=|=|=|=|}
  $A$ & AUV Frontal Area & Design Variable \\
 $l$ & AUV Length & Design Variable \\
 $u$ & AUV Control Fidelity & Design Variable \\
 $e$ & Relative Dock Entry Area & Design Variable \\
 $\eta$ & Docking Tolerance & Design Variable \\
 $h$ & Hydrodynamic Loss & Objective \\ 
 $c$ & Monetary Cost & Objective \\ 
 $d$ & Docking Reliability & Objective \\ 
 $v$ & System Versatility & Objective\\
 $p$ & Hydrodynamic Loss Weight & Minimization Parameter\\
 $q$ & Cost Weight & Minimization Parameter\\
 $r$ & Docking Success Weight & Maximization Parameter\\
 $s$ & System Versatility Weight & Maximization Parameter\\
\hline
\end{tabular}
\end{center}
\label{table:masterchart}
\end{table}

\begin{figure*}
	\centering
	\includegraphics[width=0.8\linewidth]{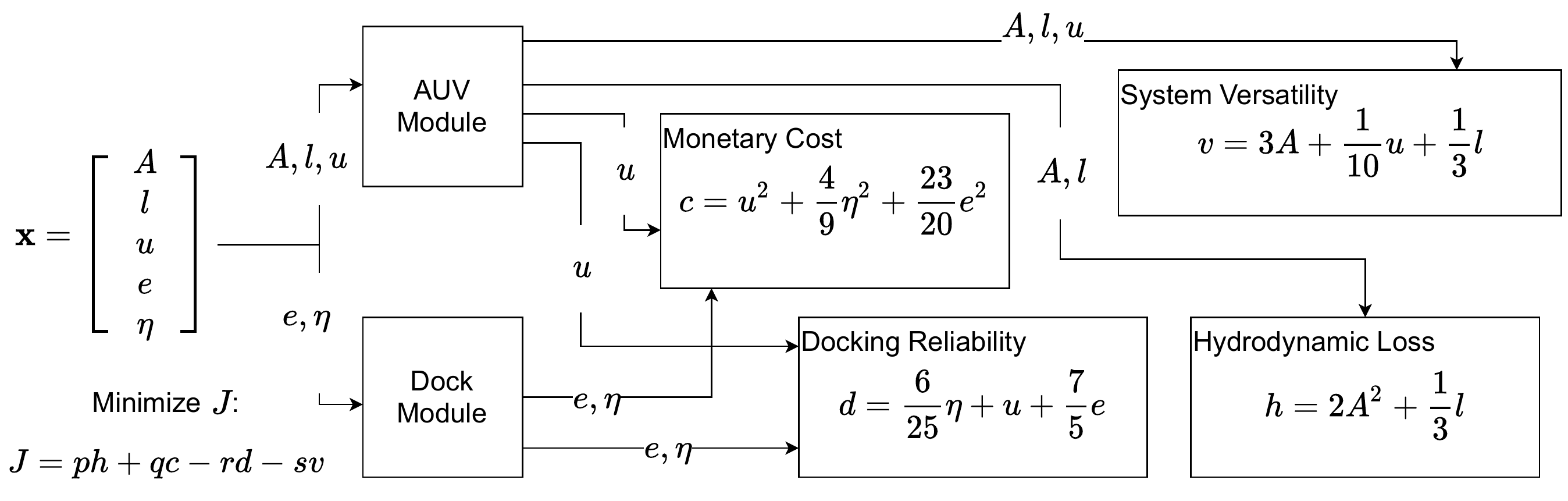}
	\vspace{-0.1in}
	\caption{Illustration of an example of implementing the proposed MDO framework for co-design optimization for an AUV docking system. The optimization variable vector consists of terms that dictates the morphology of both the AUV and its docking station. Optimization objectives are quantified based on heuristic approximations given our best knowledge of the existing systems and simplified relationship among the relevant underlying physics principles. The total optimization cost function $J$ is constructed as a weighted sum of all the individual objective terms.}
	\label{fig:model}
	\vspace{-0.0in}
\end{figure*}

To build the MDO setup, we model the set of optimization objectives (\ref{Eq.J}) as functions of the co-design parameters (\ref{Eq.X}) for the AUV-dock system. Each of these objectives is then weighted by its corresponding term inside the weight vector \eqref{eq:weight_vector} and combined together to form a final optimization function $J$. Fig.~\ref{fig:model} illustrates an example of design objectives as functions of the design variables, with the coefficients that normalize and weigh each objective term. The models seen in Fig.~\ref{fig:model} attempt to capture past expertise embedded in the body of available designs and can be adapted to contain more detail in the models or more objectives. 

\subsection{Optimization Variables} \label{des}
\begin{figure}[]
	\centering
	\includegraphics[trim={20mm 0 20mm 0}, clip, width=1\linewidth]{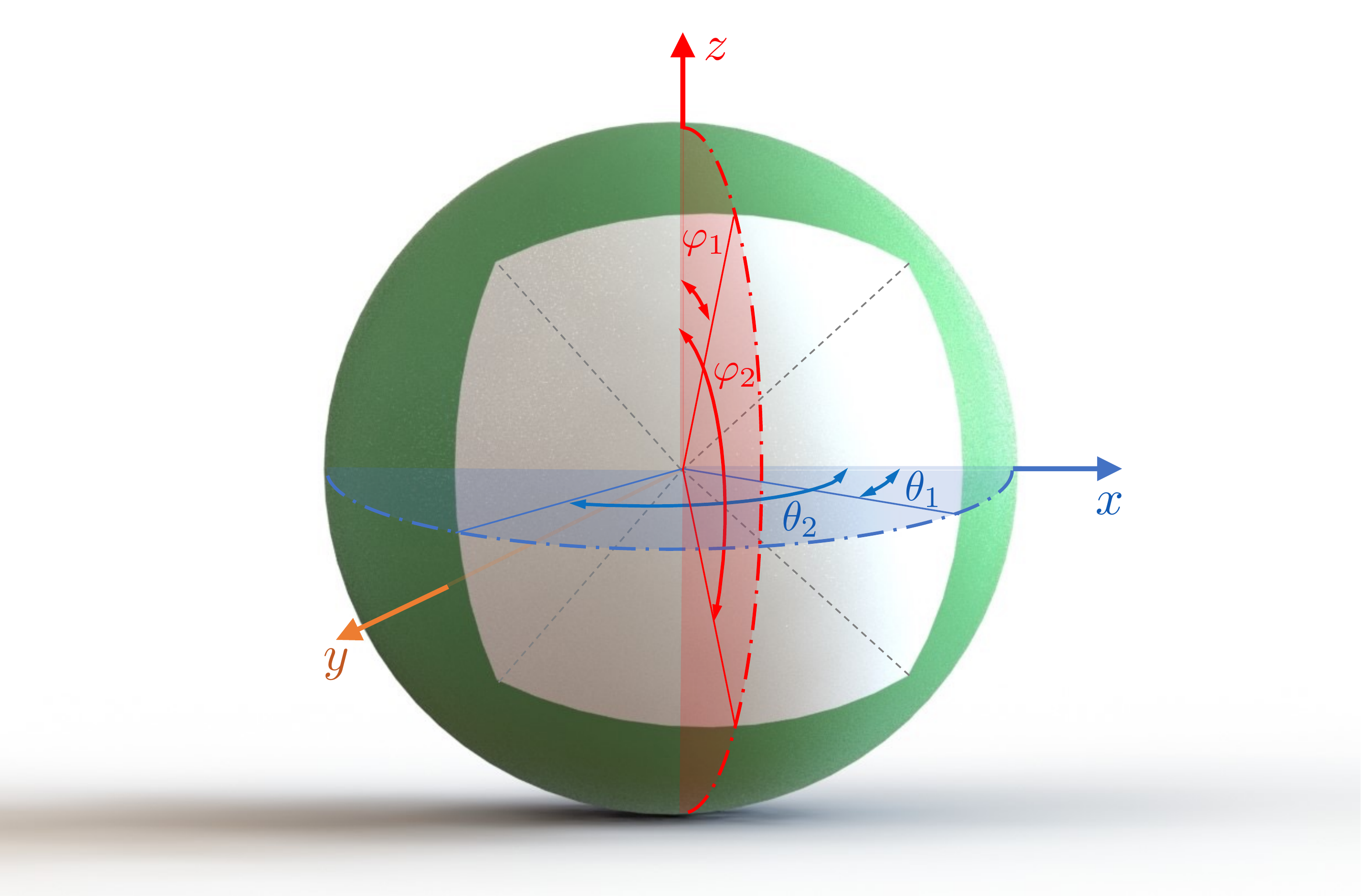}
	\vspace{-0.2in}
	\caption{A visualization of the dimensions used in defining the relative dock entry area variable, $e$. This variable is the ratio between surface area of the white portion, representing the dock entrance area, and the total surface area of a unit sphere. The relative dock entrance area can be calculated based on the values of its azimuthal angle span, $[\theta_1, \theta_2]$, and polar angle span, $[\phi_1, \phi_2]$.}
	\label{fig:sphere}
	\vspace{-0.0in}
\end{figure}

\paragraph{AUV frontal area ($A$)} The reference area of the AUV is one of the primary design variables to be considered when designing an AUV-dock system. Here, we consider the largest enclosing surface of the frontal area as the reference area, and the specific geometry of this area is not considered. The frontal area can be defined as the cross sectional area of the AUV when you are looking at the fore of the vehicle straight on. This is a similar definition that is used in most equations for cross sectional drag. The area has units of distance squared. For applications where the AUV does not enter the docking position head-on, the reference area can be taken as the corresponding  cross-section perpendicular to the principal direction of travel during docking. For instance, the largest horizontal cross-section should be used as the reference area for landing-type docking systems. 
\paragraph{AUV length ($l$)} The length of the AUV, denoted by $l$, is the body length of the AUV in the longitudinal direction. This length has a user specified unit of distance and correspond to the value that would be used to calculate skin friction. For applications where special surface properties of the AUV is considered, the vehicle length can play a non-trivial role in its hydrodynamic efficiency. Depending on the mechanism and distribution of thrusters, $l$ can also affects the mobility and agility of the vehicle during coupled maneuvering.  
\paragraph{AUV control fidelity ($u$)} We define the AUV control fidelity as a non-dimensional variable that describes the overall quality, authority, and precision of the AUV's control system including both actuation and decision making. This is modeled by two parts. First, is the number of degrees of freedom (DOF), $c_\text{DOF}$, in which the vehicle has independent control. The maximum number of DOF that the vehicle can control is six, so we describe the \textit{control authority} as
\begin{equation}
    \alpha_c = \frac{c_\text{DOF}}{6},
\end{equation}
where $c_\text{DOF}$ $\in [1,2,3,4,5,6]$. 
The second element of the control fidelity is \textit{control accuracy}, which is quantified by the relative motion control error with respect to the characteristic dimension of the vehicle. Here, we model control accuracy based on the standard deviation of the linear motion control error $\sigma_c$ \cite{ThrunS:05a} and $A$ such that
\begin{equation}
    \delta_c = \text{sat}\left(\frac{\sqrt{A}}{\sigma_c}\right),
\end{equation}
where $\text{sat}(x)$ is a saturation function that equals to $1$ for $x\geq 1$ to ensure $\delta_c \in (0, 1]$. 
The control fidelity can be defined as 
\begin{equation}
    u=\frac{w_1\alpha_c+w_2\delta_c}{w_1+w_2},
\end{equation}
where $w_1$ and $w_2$ are user-specified weighting coefficients to reflect the design prioritization between control authority and control accuracy. 
\paragraph{Relative dock entry area ($e$)}
To quantify the design space of dock morphology, we use the relative dock entry area from which the AUV can approach the docking station.
This can be seen in Fig. \ref{fig:sphere} as the portion of the surface area of the white region out of the total surface area of the sphere, which represents all the potential entry directions. 
As the entry area increases, so does the number of possible entry trajectories and maneuvers, thus increasing the likelihood of finding a successful control path \cite{WallenJ:19b}. To calculate $e$,
we first define a unit sphere centered at the entrance of the docking station. Then the entry area can be found as a function of the azimuthal angle $\theta$ and the polar angle $\phi$
\begin{equation}
    \int_{\theta_1}^{\theta_2}\int_{\phi_1}^{\phi_2}\sin\phi\ d\phi\ d\theta = |\cos\phi_2-\cos\phi_1|(\theta_2-\theta_1).
    \label{eq:entry_area}
\end{equation}
The ranges for $\theta$ and $\phi$ are denoted in Fig.~\ref{fig:sphere}.
Then the relative dock entry area can be found by normalizing the entry area \eqref{eq:entry_area} by the surface area of the unit sphere, $4\pi$, giving
\begin{equation}
   e = \frac{|\cos\phi_2-\cos\phi_1|(\theta_2-\theta_1)}{4 \pi } \in [0,1].
\end{equation}

\begin{figure}
	\centering
	\includegraphics[width=0.5\linewidth]{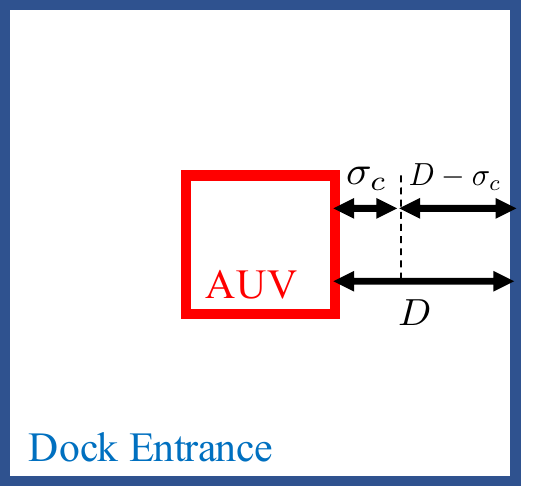}
	\vspace{-0.0in}
	\caption{A simplified visualization of a possible way to define docking tolerance based on single-side clearance, $D$, and one standard deviation of the AUV's linear motion control error, $\sigma_c$. This definition assumes that the reference surface of the AUV has to be completely inside the dock entrance are to ensure successful docking.}
	\label{fig:tolerance}
	\vspace{-0.0in}
\end{figure}

\paragraph{Docking tolerance ($\eta$)}
The last design variable we consider in this work is aimed at quantifying the amount of docking tolerance during dock entry, which is dependent on the size of capture surface of the dock and the motion control error of the AUV. In another word, it defines how tolerant the dock is to vehicle position error in order to `capture' the AUV into a docked state.  Fig.~\ref{fig:tolerance} gives a visualization of our definition of docking tolerance. The value $D = \sigma_c+\delta_d$ denotes the single-side clearance between the AUV and the boundary of the dock entry when the AUV docks at the ideal entry point. The vehicle will not dock successfully when $D < 0$. With control error of one standard deviation, this leaves a remaining clearance of $D-\sigma_c$. We quantify docking tolerance with the normalized remaining clearance, i.e.,
\begin{equation}
    \eta = \text{sat}\left(\frac{D-\sigma_c}{\sigma_c}\right) \in [0,1].
\end{equation}

\subsection{Optimization Objectives} \label{opt} 
\paragraph{Hydrodynamic loss ($h$)}
In an effort to maximize the hydrodynamic efficiency of the AUV, i.e., minimizing its hydrodynamic loss, we model an optimization objective that is dependent on the frontal area of the AUV and the length of the AUV. These two variables contribute to the form drag/added mass and skin friction, respectively. The term for the form drag is considered as a quadratic function of the reference frontal area to reflect its larger contribution to the overall hydrodynamic effect of the AUV compared to its length \cite{liggetJ1994}. While we do not delve into the intricacies of hydrodynamics in this work, our optimization frameworks is readily compatible with other high-fidelity models for hydrodynamic losses. 
\paragraph{Monetary cost ($c$)}
The second optimization objective used in this problem setup is the monetary cost of the system, including both the AUV and the dock. Contributing to this cost term are the AUV control fidelity, relative dock entry area, and docking tolerance. In practice, the relation between the $c$ and its dependent variables varies with several economical factors such as the cost decreases of relevant technologies as they mature, the level of supply-demand imbalance for parts and components, etc. The user can obtain an application-specific composition of the dependent variables of $c$. The proper weighting among these variables can be obtained heuristically by analyzing scaling of the build cost with respect to each of the variables at the time of design. In our case, we consider $c$ as a quadratic function of its dependent variables as a proper estimate based on the current market and our recent experiences in budgeting for AUV-dock systems. 
\paragraph{Docking reliability ($d$)}
An important requirement of any AUV docking system is its reliability in successfully docking the AUV at the dock such that the dock can properly provide the necessary power and communication access to the vehicle. As a critical part of the system, docking reliability ultimately defines the success of the mission and vehicle operations~\cite{PageB}. To model docking reliability, we include variables including the AUV control fidelity, relative dock entry area, and docking tolerance. We use a linear relationship to capture the dependency of $d$ on its dependent variables. 
\paragraph{System versatility ($v$)}
While all the other optimization variables would create a more ideal system, the resulting system may be challenging to work on due to its small internal volume or limited maneuverability to attend to real-world applications. For example, the limited number of sensors able to fit on board may end up creating an inconvenience for the user \cite{DeutschC2018}\cite{eichhornM2018}. To ensure an efficient but also versatile system, we introduce a system versatility optimization objective that regularizes the dimensions and the control fidelity of the resulting AUV.

\section{Results and Analysis}
The optimization cost function is built from the optimization objectives discussed before and is expressed as
\begin{equation}
   J = \mathbf{w}\mathbf{j}= ph + qc - rd - sv.
    \label{Optimization Equation}
\end{equation}
This cost function casts the underlying physical principals and design heuristics into a multi-objective optimization problem, resulting in an optimal design. Here, the minimization of $J$ is conducted in MATLAB utilizing the `\textit{fmincon}' function \cite{LeLuongH2019}. Using the interior point optimization algorithm \cite{cdi_springer_books_10_1007_b100325}, the resulting vector of design variables is found. To initiate the process, the optimization weights ($p, q, -r, -s$) are user defined and initial guesses for the optimization variables, $\mathbf{x}_\text{init}$, are provided to the optimization solver. This process results in a set of optimal design parameters, $\mathbf{x}^*$, that can be used to inform the AUV-dock system design. 

\subsection{Test Setup}
Table \ref{table:bounds} lists the bounds for the optimization variables in our MDO setup, constraining the search space for the optimal solution. These bounds can be adapted to meet a user's specific needs. 
The following initial guess was used in all of our test cases: 
\begin{equation*}
\mathbf{x}_\text{init}= 
\begin{bmatrix}
0.03\ m^2,\ 1.5\ m,\ 0.5,\ 0.5,\ 0.5 
\end{bmatrix}^T. 
\label{Eq.x}
\end{equation*}

\begin{table}
\caption{Optimization Variable Bounds}
\begin{center}
\begin{tabular}{|c|c|c|} 
 \hline
\textbf{ Lower Bound} & \textbf{Variable} & \textbf{Upper Bound}  \\ 
  \hhline{|=|=|=|}
 $0.01\ m^2$ & $A$ & $1\ m^2$ \\
 $0.5\ m$ & $l$ & $3\ m$  \\
 $0.083$ & $u$ & $1$ \\
 $0.177$ & $e$ & $0.855$  \\
 $0$ & $\eta$ & $1$  \\
\hline
\end{tabular}
\end{center}
\label{table:bounds}
\end{table}

To further constrain the search space, nonlinear inequality constraints were added. The AUV volume constraint  
\begin{equation*}
    0.025\ m^3 \leq A*l
\end{equation*}
corresponds to the physical volume of the system and ensures that the system will not be excessively small resulting in a solution, which may be challenging or impossible to implement in practice. The docking tolerance constraint 
\begin{equation*}
    1.5\ m^{-2} \leq \frac{\eta}{A}
\end{equation*}
specifies that as the overall size of the AUV increases, the docking tolerance should increase accordingly to account for the increasing scale of the docking process. 
 
\subsection{Test Results}
The MDO framework and implementation that has been described up to this point has been applied to three different cases. First, the design intent was determined that corresponds to a real world design problem. Then the user defined weights were set to represent the design problem. For each case, the optimization weights, optimal design vector, and closely corresponding real world examples are presented and discussed. 

\subsubsection{Case 1 - General Design Intent}

In this first test, a general design intent was investigated. This design intent corresponds to the case where each of the optimization objectives is considered equally important. To represent this design intent in the MDO framework, the optimization weights are set as 
\begin{equation*}
\mathbf{w}_\text{general}=\begin{bmatrix}
1,\ 1,\ -1,\ -1
\end{bmatrix},
\end{equation*}
resulting in an optimal design vector of 
\begin{equation*}
\textbf{x}^{\ast}_\text{general} =
\begin{bmatrix}
0.506\ m^2,\ 2.1\ m,\ 0.55,\ 0.61,\ 0.76
\end{bmatrix}^T.
\label{eq:General}
\end{equation*}

\begin{figure}[h]
	\centering
	\includegraphics[width=1\linewidth]{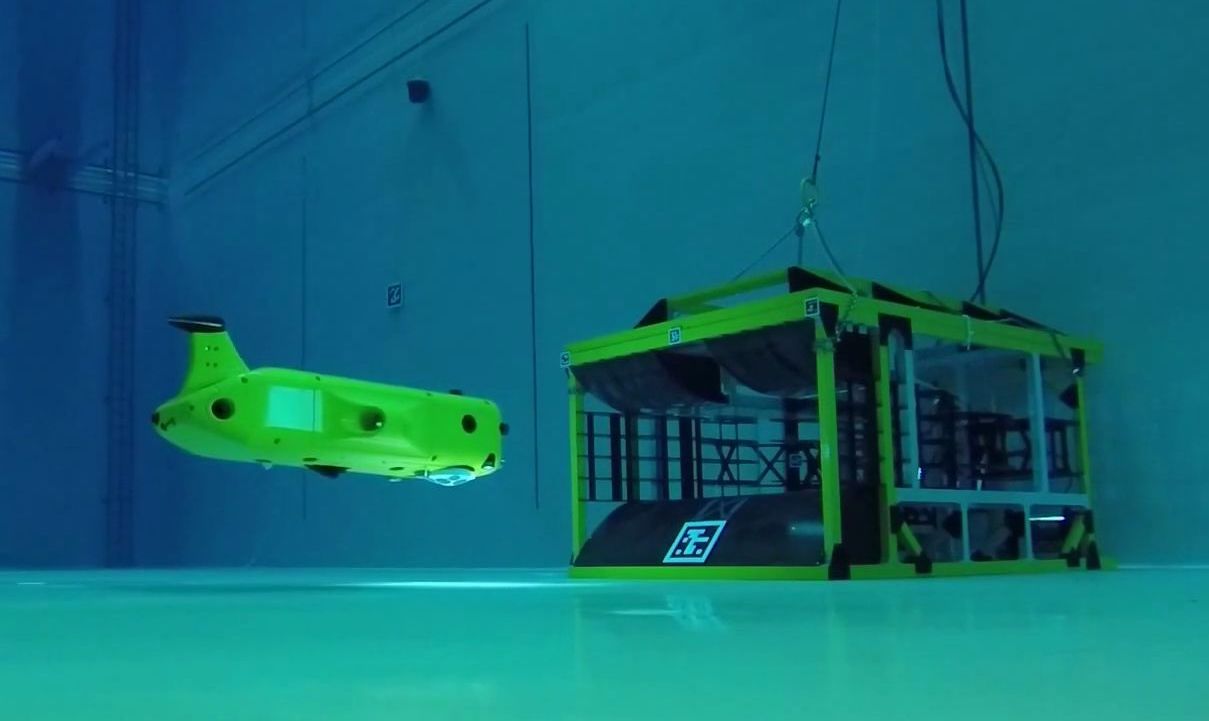}
	\vspace{-0.2in}
	\caption{The Flatfish AUV~\cite{AlbiezJ:15a} and its docking station are an example of general-purpose AUV docking system and a real-world example of an optimal design for the general design intent case. (Image Credit: German Research Center for Artificial Intelligence)}
	\label{fig:flatfish}
	\vspace{-0.0in}
\end{figure}

To validate the optimization solution, real-world design parallels can be made. One such system with similar design variables can be seen in the Flatfish AUV~\cite{AlbiezJ:15a} and docking station in a project jointly undertaken by the German Research Center for Artificial Intelligence (DFKI), the Brazilian Institute of Robotics (BIR) and Shell (Fig.~\ref{fig:flatfish}). The Flatfish and other analogous systems can be described as general-purpose AUV-dock systems. They are hydrodynamically efficient and employ 3-6 degrees of freedom of control. Additionally, this style of docking station resembles an airplane hanger accompanied by mechanical guidance, ultimately allowing for a large entry area (approximately half of a sphere ). It can also be seen that the AUV dimensions are consistent with those found in our MDO solution. 

\subsubsection{Case 2 - Low-cost Design Intent}
The second design scenario considered was the low-cost design intent. This design intent is often seen in scenarios where the cost of the overall system is a main factor in making design choices. This could be the case for small companies or academic groups with limited resources or for deploying a scalable networked system consisting of multiple AUVs and docking stations. This intent is characterized by a larger weight for the monetary cost term in $J$. The weight vector in this case was set as
\begin{equation*}
\mathbf{w}_\text{low-cost}=\begin{bmatrix}
1,\ 2,\ -1,\ -1
\end{bmatrix},
\end{equation*}
resulting in an optimal design vector of
\begin{equation*}
\textbf{x}^{\ast}_\text{low-cost} = 
\begin{bmatrix}
0.38\ m^2,\ 2.11\ m,\ 0.275,\ 0.30,\ 0.57
\end{bmatrix}^T.
\end{equation*}

\begin{figure}
	\centering
	\includegraphics[width=1\linewidth]{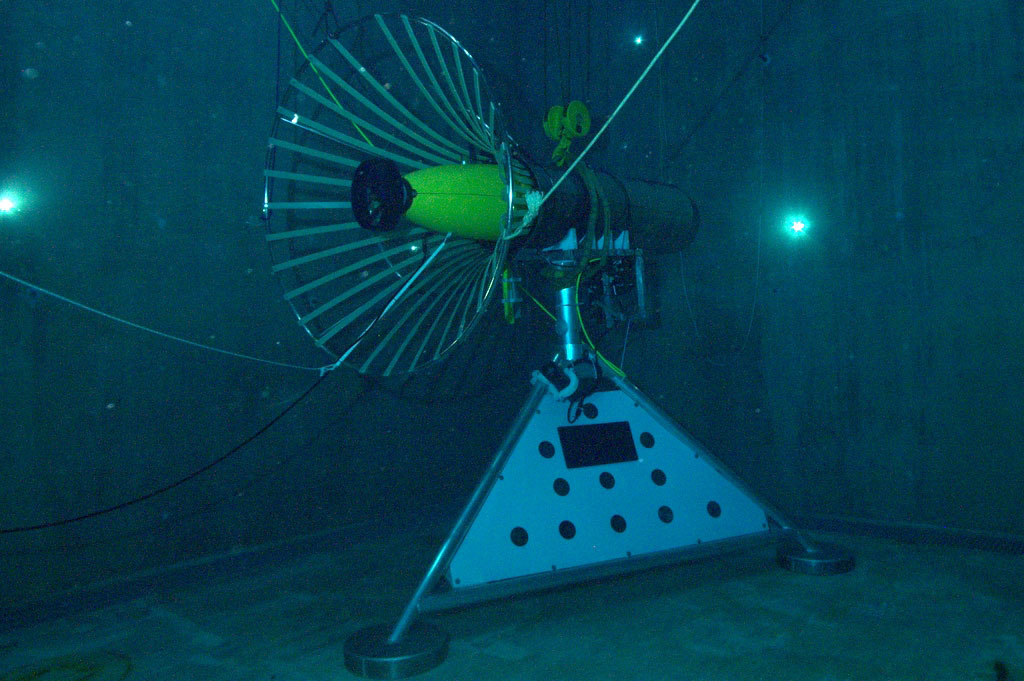}
	\vspace{-0.2in}
	\caption{The 21-inch torpedo-form AUV with a funnel-type docking station developed by the Monterey Bay Aquarium Research Institute (MBARI)~\cite{HobsonB:07a} is consistent with the optimal design parameters found by our MDO for a low-cost AUV docking system. (Image Credit: MBARI)}
	\label{fig:mbari}
	\vspace{-0.0in}
\end{figure}

\begin{figure*}
	\centering
	\includegraphics[width=1\linewidth]{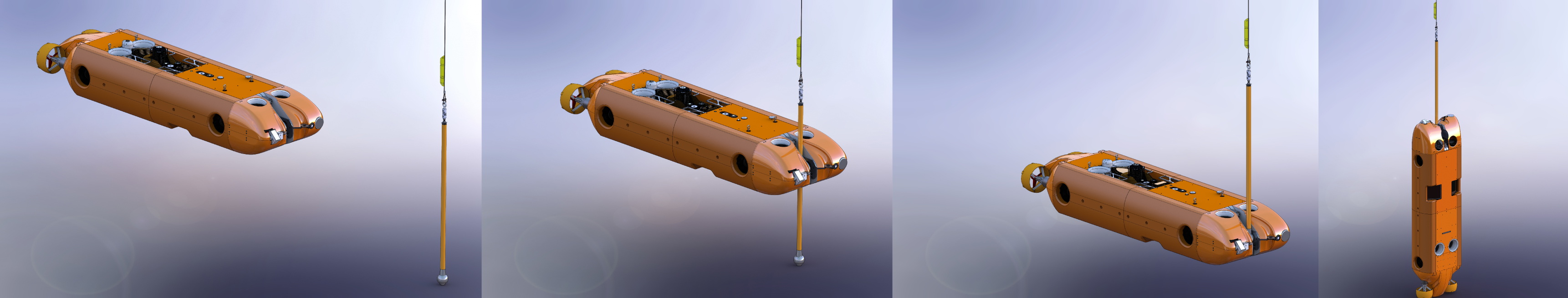}
	\vspace{-0.2in}
	\caption{The ARTEMIS AUV and lure docking system \cite{kimballP2018} is an under ice surveying AUV that matches the optimal design found by the proposed multidisciplinary optimization framework for the survey mission design intent. (Image Credit: Stone Aerospace)}
	\label{fig:artemis}
	\vspace{-0.0in}
\end{figure*}

With this result, we can draw a parallel to an AUV docking station system archetype that has similar design variable as that seen in Fig.~\ref{fig:mbari}. The 21-inch AUV and docking station developed by the Monterey Bay Aquarium and Research Institute (MBARI)~\cite{HobsonB:07a}. This system consists of a torpedo-form vehicle with a funnel type docking station. It can be seen that the AUV dimensions are consistent with the optimal design results. Compared with the result from the general design intent, the MBARI system has a smaller relative dock entry area and smaller docking tolerance, which is mostly due to the lower control fidelity of the AUV. Although it is difficult to argue that one archetype is cheaper than another since monetary cost depends on quality of components and the time of development, the MDO result suggests that this archetype will generally be more cost effective if that is the user's intent. 

\subsubsection{Case 3 - Survey Mission Design Intent}
Lastly, we look at the case where the user desires an accurate and efficient surveying AUV that can be equipped with various payload packages and docked for charging and data transfer. This design intent is often seen in the design of a resident AUV system~\cite{ManalangDA:16a,MatsudaT:19a,SongZ2020}. In our MDO framework, this design intent is expressed as 
\begin{equation*}
\mathbf{w}_\text{survey}=
\begin{bmatrix}
2,\ 1,\ -1.2,\ -2
\end{bmatrix}.
\end{equation*}
After performing the optimization process, the resulting design variables vector is
\begin{equation*}
\mathbf{x}^{\ast}_\text{survey} =
\begin{bmatrix}
0.604\ m^2,\ 2.09\ m,\ 0.70,\ 0.73,\ 0.91
\end{bmatrix}^T.
\end{equation*}

These design parameters indicate a surveying AUV docking system that is similar to the ARTEMIS AUV, a 20-km-range hover-capable hybrid AUV, and its docking system developed by Stone Aerospace as part of the Sub-Ice Marine Planetary Analog Ecosystems (SIMPLE) project, supported by NASA Astrobiology \cite{kimballP2018}. The Artemis is an under ice surveying AUV that has a lure type docking system. This docking method is approachable from 360 degrees in the yaw direction, yielding a very high docking entry area. Additionally, the form factor of the AUV is versatile and can be equipped with various payloads. The dimensions of the ARTEMIS AUV match the optimization result well. 

\subsection{Evaluation of Results}
The three cases that have been investigated and reported present a preliminary validation of this MDO system. The purpose of this investigation has been to show that an analytical design optimization tool can be used to capture heuristic design decisions that have been made by subject-matter experts over the past few decades. The results presented and their corresponding real-world systems are not perfect interpretations of the optimal AUV docking system designs. Nonetheless, our findings show that the general design intents can be captured by MDO and design optimization should be considered simultaneously for both the AUV and its docking station. It is also promising for our method to become a useful tool for students and educators who are not necessarily experts in the area of ocean engineering and AUV design.  




\section{Conclusion and Future Work}
A multi-objective design optimization framework has been presented and implemented for the co-design of autonomous underwater vehicle (AUV) docking station systems. The motivation behind AUV-dock co-design was discussed and served as a foundation for the formulation of the optimization problem. After defining the design statement, design objective, and design parameters, an optimization simulation for AUV-dock co-design was formulated. The proposed co-design optimization framework results in an optimal design solution for both the AUV and its docking station by minimizing a user-defined objective function. This framework quantifies user design intents and constrains the search of the optimal solution within reasonable bounds. Three test cases were presented to showcase the capability of the proposed co-design framework in capturing the design intents for a general-purpose AUV, a low-cost AUV, and a survey-class AUV. Existing systems that are consistent with the resulting optimal co-design parameters are identified to discuss the validity and limitation of the proposed method. While this study does not claim that ideal real-world systems should perfectly match the optimization results, this MDO framework is an useful analytical design tool that can be used to guide concept selection and aid in the design optimization of AUV docking systems.

The proposed co-design method can be improved to better incorporate high-level design intents that affect both the AUV and the docking station. First, the simplified modeling used to describe the physical and design components in our cost function has left room for more details to be added in capturing more aspects of the problem such as the complexity trade-off between the AUV and the dock, system longevity, depth rating, etc. Another element that would make this co-design framework more versatile would be the ability to generate the optimal docking station design parameters given an AUV archetype as input. This would aid in the development of docking stations for existing AUVs and vice versa. Lastly, future work should involve the development of an automated parametric modeling system that will take as input the optimized design variables and output a three dimensional model of the optimal archetype. This would be a useful tool in the brainstorming and design process. The presented work coupled with the possible future work show promising potential for analytical co-design of AUV docking station systems.

\section*{Acknowledgment}
JW and MJ would like to thank Dr.~Reza Ghorbani for the valuable discussions about MDO. This work was supported by the U.S.~Department of Energy (DOE) under project National Marine Renewable Energy Center Infrastructure Upgrades. JW and ZS would also like to acknowledge funding support from the U.S.~National Science Foundation (NSF CISE/IIS–2024928).

\bibliographystyle{IEEEtran}
\bibliography{RAUV,ref,ran}

\end{document}